\documentclass[conference]{IEEEtran}

\usepackage{cite}
\usepackage{amsmath,amssymb,amsfonts}
\usepackage{algorithmic}
\usepackage{graphicx}
\usepackage{textcomp}
\usepackage{xcolor}

\usepackage{booktabs}
\usepackage{dirtytalk}
\usepackage{latexsym} 
\usepackage{url}
\usepackage{multicol}
\usepackage{enumitem}
\usepackage[super]{nth}
\usepackage{thmtools}
\usepackage{stfloats}

\usepackage{booktabs}

\usepackage{verbatim} 
\newcommand{\minus}{\scalebox{0.65}[1.0]{$-$}}
\newcommand{\minusminus}{\minus 1/\minus 1 }
\newcommand{\minusminusnospace}{\minus 1/\minus 1}
\newcommand{\minusminustwo}{\minus 2/\minus 2 }
\newcommand{\minusminustwonospace}{\minus 2/\minus 2}

\newtheorem{thm}{Theorem}

\newtheorem{conj}{Conjecture}
\newtheorem{open}[conj]{Open Problem}

\def\BibTeX{{\rm B\kern-.05em{\sc i\kern-.025em b}\kern-.08em
    T\kern-.1667em\lower.7ex\hbox{E}\kern-.125emX}}
  
\begin{document}

\title{\textit{Magic: The Gathering} is Turing Complete}

\author{
\IEEEauthorblockN{Alex Churchill}
\IEEEauthorblockA{Independent Researcher\\
  Cambridge, United Kingdom\\
  \texttt{alex.churchill@cantab.net}}
\and
\IEEEauthorblockN{Stella Biderman}
\IEEEauthorblockA{Georgia Institute of Technology\\
  Atlanta, United States of America\\
  \texttt{stellabiderman@gatech.edu}}
\and
\IEEEauthorblockN{Austin Herrick}
\IEEEauthorblockA{University of Pennsylvania\\
  Philadelphia, United States of America\\
  \texttt{aherrick@wharton.upenn.edu}}
}
\maketitle

\begin{abstract}
\textit{Magic: The Gathering} is a popular and famously complicated trading card game about magical combat. In this paper we show that optimal play in real-world \textit{Magic} is at least as hard as the Halting Problem, solving a problem that has been open for a decade \cite{dh:games-puzzles-computation,at:frontier}. To do this, we present a methodology for embedding an arbitrary Turing machine into a game of \textit{Magic} such that the first player is guaranteed to win the game if and only if the Turing machine halts. Our result applies to  how  real \textit{Magic} is  played, can be achieved using standard-size tournament-legal decks, and does not rely on stochasticity or hidden information. Our result is also highly unusual in that all moves of both players are forced in the construction. This shows that even recognising who will win a game in which neither player has a non-trivial decision to make for the rest of the game is undecidable. We conclude with a discussion of the implications for a unified computational theory of games and remarks about the playability of such a board in a tournament setting.
\end{abstract}

\section{Introduction}
\textit{Magic: The Gathering} (also known as \textit{Magic}) is a popular trading card game owned by Wizards of the Coast. Formally, it is a two-player zero-sum stochastic card game with imperfect information, putting it in the same category as games like poker and hearts. Unlike those games, players design their own custom decks out of a card-pool of over 20,000 cards. \textit{Magic}'s multifaceted strategy has made it a popular topic in artificial intelligence research.

In this paper, we examine \textit{Magic: The Gathering} from the point of view of algorithmic game theory, looking at the computational complexity of evaluating who will win a game. As most games have finite limits on their complexity (such as the size of a game board) most research in algorithmic game theory of real-world games has primarily looked at generalisations of commonly played games rather than the real-world versions of the games. A few real-world games have been found to have non-trivial complexity, including \textit{Dots-and-Boxes}, \textit{Jenga} and \textit{Tetris} \cite{dh:cgt}. We believe that no real-world game is known to be harder than NP previous to this work.

Even when looking at generalised games, very few examples of undecidable games are known. On an abstract level, the Team Computation Game \cite{dh:contraint-logic} shows that some games can be undecidable, if they are a particular kind of team game with imperfect information. The authors also present an equivalent construction in their Constraint Logic framework that was used by Coulombe and Lynch (2018) \cite{cl:smash} to show that some video games, including \textit{Super Smash Bros Melee} and \textit{Mario Kart}, have undecidable generalisations. Constraint Logic is a highly successful and highly flexible framework for modelling games as computations.

The core of this paper is the construction presented in Section \ref{full}: a universal Turing machine embedded into a game of \textit{Magic: The Gathering}. As we can arrange for the victor of the game to be determined by the halting behaviour of the Turing machine, this construction establishes the following theorem:

\begin{thm}\label{thm:0player}
Determining the outcome of a game of \textit{Magic: The Gathering} in which all remaining moves are forced is undecidable.
\end{thm}

\subsection{Previous Work}
Prior to this work, no undecidable real games were known to exist. Demaine and Hearn (2009) \cite{dh:games-puzzles-computation} note that almost every real-world game is trivially decidable, as they produce game trees with only computable paths. They further note that Rengo Kriegspiel\footnote{Rengo Kriegspiel is a combination of two variations on Go: Rengo, in which two players play on a team alternating turns, and Shadow Go, in which players are only able to see their own moves.} is ``a game humans play that is not obviously decidable; we are not aware of any other such game.'' It is conjectured by Auger and Teytaud (2012) \cite{at:frontier} that Rengo Kriegspiel is in fact undecidable, and it is posed as an open problem to demonstrate any real game that is undecidable.

The approach of embedding a Turing machine inside a game directly is generally not considered to be feasible for real-world games \cite{dh:games-puzzles-computation}. Although some open-world sandbox games such as Minecraft and Dwarf Fortress can support the construction of Turing machines, those machines have no strategic relevance and those games are deliberately designed to support large-scale simulation. In contrast, leading formal theory of strategic games claims that the unbounded memory required to simulate a Turing machine entirely in a game would be a violation of the very nature of a game \cite{dh:contraint-logic}.

The computational complexity of \textit{Magic: The Gathering} in has been studied previously by several authors. Our work is inspired by \cite{c:old-tm}, in which it was shown that four-player \textit{Magic} can simulate a Turing machine under certain assumptions about player behaviour. In that work, Churchill conjectures that these limitations can be removed and preliminary work along those lines is discussed in \cite{c:forum}. The computational complexity of checking the legality of a particular decision in \textit{Magic} (blocking) is investigated in \cite{ci:blocking} and is found to be coNP-complete. There have also been a number of papers investigating algorithmic and artificial intelligence approaches to playing \textit{Magic}, including Ward and Cowling (2009) \cite{wc:mcmc}, Cowling et al. (2012) \cite{wcp:mcmc}, and Esche (2018) \cite{e:thesis}. Esche (2018) briefly considers the theoretical computational complexity of \textit{Magic} and states an open problem that has a positive answer only if \textit{Magic} end-games are decidable.

\subsection{Our Contribution}
This paper completes the project started by Churchill \cite{c:old-tm} and continued by Churchill et al. \cite{c:forum} of embedding a universal Turing machine in \textit{Magic: The Gathering} such that determining the outcome of the game is equivalent to determining the halting of the Turing machine. This is the first result showing that there exists a real-world game for which determining the winning strategy is non-computable, answering an open question of Demaine and Hearn \cite{dh:games-puzzles-computation} and Auger and Teytaud \cite{at:frontier} in the positive. This result, combined with Rice's Theorem \cite{rice}, also answers an open problem from Esche \cite{e:thesis} in the negative by showing that the equivalence of two strategies for playing \textit{Magic} is undecidable.

This result raises important foundational questions about the nature of a game itself. As we have already discussed, the leading formal theory of games holds that this construction is unreasonable, if not impossible, and so a reconsideration of those assumptions is called for. In section \ref{theory-of-gaming} we discuss additional foundational assumptions of Constraint Logic that \textit{Magic: The Gathering} violates, and present our interpretation of the implications for a unified theory of games.

\subsection{Overview}
The paper is structured as follows. In Section \ref{prelim} we provide background information on this work, including previous work on \textit{Magic} Turing machines. In Section \ref{overview} we present a sketch of the construction and its key pieces. In Section \ref{full} we provide the full construction of a universal Turing machine embedded in a two-player game of \textit{Magic}. In Section \ref{discussion} we discuss the game-theoretic and real-world implications of our result.

\section{Preliminaries}\label{prelim}
One initial challenge with \textit{Magic: The Gathering} is the encoding of information. Some cards ask players to choose a number. Although rules for how to specify a number are not discussed in the \textit{Comprehensive Rules} \cite{rules}, convention is that players are allowed to specify numbers in any way that both players can agree to. For example, you are allowed to choose the number $2^{100}$ or $\lceil\log 177\rceil$. This presents an issue brought to our attention by Fortanely \cite{f:incomparable}. Consider the following situation: both players control \textbf{Lich}, \textbf{Transcendence}, and \textbf{Laboratory Maniac}. One player then casts \textbf{Menacing Ogre}. The net effect of this play is the ``Who Can Name the Bigger Number'' game -- whoever picks the biggest number wins on the spot. This makes identifying the next board state non-computable \cite{bk:biggest}, so we require that any numbers specified by a player must be expressed in standard binary notation.

We believe that with this restriction \textit{Magic: The Gathering} is \textit{transition-computable}, meaning that the function that maps a board state and a move to the next board state is computable\footnote{We avoid the term ``computable game'' which is more commonly used to mean that the game has a computable winning strategy.}. However, it is unclear how to prove this beyond exhaustive analysis of the over 20,000 cards in the game. We leave that question open for future work:
\begin{conj}\label{tcomp}
The function that takes a board state and a legal move and returns the next board state in \textit{Magic: The Gathering} is computable.
\end{conj}
In this conjecture we say ``a legal move'' because it is also not obvious that checking to see if a move is legal is computable. Chatterjee and Ibsen-Jensen \cite{ci:blocking} show that checking the legality of a particular kind of game action is coNP-complete, but the question has not been otherwise considered. Again, we leave this for future work:
\begin{conj}\label{checking}
There does not exist an algorithm for checking the legality of a move in \textit{Magic: The Gathering}.
\end{conj}

\subsection{Previous \textit{Magic} Turing Machines}
In \cite{c:old-tm}, the author presents a \textit{Magic: The Gathering} end-game that embeds a universal Turing machine. However, this work has a major issue: it's not quite deterministic. At several points in the simulation, players have the ability to stop the computation at any time by opting to decline to use effects that say ``may.'' For example, \textbf{Kazuul Warlord} reads ``Whenever Kazuul Warlord or another Ally enters the battlefield under your control, you may put a $+1/+1$ counter on each Ally you control.'' Declining to use this ability will interfere with the Turing machine, either causing it to stop or causing it to perform a different calculation from the one intended. The construction as given in Churchill \cite{c:old-tm} works under the assumption that all players that are given the option to do something actually do it, but as the author notes it fails without this assumption. Attempts to correct this issue are discussed in Churchill et al. \cite{c:forum}.

In this work, we solve this problem by reformulating the construction to exclusively use cards with mandatory effects. We also substantially simplify the most complicated aspect of the construction, the recording of the tape, and reduce the construction from one involving four players to one involving two, and which only places constraints on one player's deck, matching the format in which \textit{Magic} is most commonly played in the real world (two-player duels). Like the previous work, we will embed Rogozhin's (2,~18) universal Turing machine \cite{r:utms}.


\section{An Overview of the Construction}\label{overview}
In this section we give a big picture view of the Turing machine, with full details deferred to the next section. The two players in the game are named Alice and Bob.

To construct a Turing machine in \textit{Magic: The Gathering} requires three main elements: the tape which encodes the computation, the controller which determines what action to take next based on the current state and the last read cell, and the read/write head which interacts with the tape under the control of the controller.

\subsection{The Tape}
As the rules of \textit{Magic: The Gathering} do not contain any concept of geometry or adjacency, encoding the tape itself is tricky. Our solution is to have many creature tokens with carefully controlled power and toughness, with each token's power and toughness representing the distance from the head of the Turing machine. The tape to the left of the Turing machine's current read head position is represented by a series of creature tokens which all have the game colour green, while the tape to the right is represented by white tokens. Our distance-counting starts at 2, so there is one 2/2 token representing the space currently under the head of the Turing machine; a green 3/3 token represents the tape space immediately to the left of the Turing head, a green 4/4 is the space to the left of that, and so on. Rogozhin's universal Turing machine starts with the read head in the middle of the tape\cite{r:utms}.

To represent the symbols on the tape, we use creature types. We choose 18 creature types from the list of creature types in \textit{Magic} to correspond to the 18 symbols in Rogozhin's (2,~18) UTM. We can choose these creature types to begin with successive letters of the alphabet: Aetherborn, Basilisk, Cephalid, Demon, Elf, Faerie, Giant, Harpy, Illusion, Juggernaut, Kavu, Leviathan, Myr, Noggle, Orc, Pegasus, Rhino, and Sliver. For example, a green 5/5 Aetherborn token represents that the \nth{1} symbol is written on the \nth{3} cell to the left of the head, and a white 10/10 Sliver represents that the \nth{18} symbol is written on the \nth{9} cell to the right of the head. These tokens are all controlled by Bob, except the most recently created token (the space the Turing head has just left) which is controlled by Alice.

\subsection{The Controller}

Control instructions in a Turing machine are represented by a table of conditional statements of the form ``if the machine is in state $s$, and the last read cell is symbol $k$, then do such-and-such.'' Many \textit{Magic} cards have triggered abilities which can function like conditional statements. The two we shall use are \textbf{Rotlung Reanimator} (``Whenever Rotlung Reanimator or another Cleric dies, create a 2/2 black Zombie creature token'') and \textbf{Xathrid Necromancer} (``Whenever Xathrid Necromancer or another Human creature you control dies, create a tapped 2/2 black Zombie creature token''). We will use both, and the difference between tapped and untapped creature tokens will contribute to the design of the Turing machine.

Each \textbf{Rotlung Reanimator}\footnote{For now we will speak about \textbf{Rotlung Reanimator} for simplicity. Some of these will in fact be \textbf{Xathrid Necromancer}s as explained in the next section.} needs to trigger from a different state being read -- that is, a different creature type dying -- and needs to encode a different result. Fortunately, \textit{Magic} includes cards that can be used to edit the text of other cards. The card \textbf{Artificial Evolution} is uniquely powerful for our purposes, as it reads ``Change the text of target spell or permanent by replacing all instances of one creature type with another. The new creature type can't be Wall. \textit{(This effect lasts indefinitely.)}'' So we create a large number of copies of \textbf{Rotlung Reanimator} and edit each one. A similar card \textbf{Glamerdye} can be used to modify the colour words within card text.

Thus, we edit a \textbf{Rotlung Reanimator} by casting two copies of \textbf{Artificial Evolution} replacing `Cleric' with `Aetherborn' and `Zombie' with `Sliver' and one copy of \textbf{Glamerdye} to replace `black' with `white', so that this \textbf{Rotlung Reanimator} now reads ``Whenever Rotlung Reanimator or another \textit{Aetherborn} dies, create a 2/2 \textit{white} \textit{Sliver} creature token''\footnote{Throughout this paper, card text that has been modified using cards such as \textbf{Artificial Evolution} is written in italics.}. This \textbf{Rotlung Reanimator} now encodes the first rule of the $q_1$ program of the (2,~18) UTM: ``When reading symbol 1 in state A, write symbol 18 and move left.'' The Aetherborn creature token represents symbol 1, the Sliver creature token represents symbol 18, and the fact that the token is white leads to processing that will cause the head to move left.

We similarly have seventeen more \textbf{Rotlung Reanimator}s encoding the rest of the $q_1$ program from \cite{r:utms}. Between them they say:

\begin{enumerate}
    \item Whenever an \textit{Aetherborn} dies, create a 2/2 \textit{white Sliver}.
    \item Whenever a \textit{Basilisk} dies, create a 2/2 \textit{green Elf}.
    \item[ ] Whenever a $\ldots$ dies, create a 2/2 $\ldots$
    \item[18)] Whenever a \textit{Sliver} dies, create a 2/2 \textit{green Cephalid}.
\end{enumerate}
See Table \ref{tab:program} for the full encoding of the program.

\subsection{The Read/Write Head}
The operation ``read the current cell of the tape'' is represented in-game by forcing Alice to cast \textbf{Infest} to give all creatures \minusminustwonospace. This causes the unique token with 2 toughness to die. It had a colour (green or white) which is irrelevant, and a creature type which corresponds to the symbol written on that cell. That creature type is noticed by a \textbf{Rotlung Reanimator}, which has a triggered ability that is used to carry out the logic encoded in the head of the Turing machine. It produces a new 2/2 token, containing the information written to the cell that was just read.

The Turing machine then moves either left or right and modifies the tokens to keep the tape in order by adding \mbox{$+1/+1$} counters to all tokens on one side of the head and \minusminus counters to all tokens on the other side. 
Moving left or right will be accomplished by casting first \textbf{Cleansing Beam} and then \textbf{Soul Snuffers}.

\subsection{Adding a Second State}
Everything described so far outlines the operation of one state of the Turing machine. However, our Turing machine requires two states. To accomplish this, we leverage \textit{phasing}: an object with phasing can `phase in' or `phase out', and while it's phased out, it's treated as though it doesn't exist. We can grant phasing to our \textbf{Rotlung Reanimator}s using the enchantment \textbf{Cloak of Invisibility} (``Enchanted creature has phasing and can't be blocked except by Walls'') and create a second set of \textbf{Rotlung Reanimator}s to encode the program $q_2$. At the moment we read the current cell, exactly one set of \textbf{Rotlung Reanimator}s will be phased in.

Objects with phasing phase in or out at the beginning of their controller's turn, effectively toggling between two states. Accordingly we will arrange for the turn cycle to last 4 turns for each player when no state change occurs, but just 3 turns when we need to change state.

\section{The Full Construction}\label{full}
Now we will provide the full construction of the \textit{Magic: The Gathering} Turing machine and walk through a computational step. The outline of one step of the computation is as follows (Bob's turns are omitted as nothing happens during them):

\begin{itemize}[itemsep=4pt,parsep=2pt] 
    \item[T1] Alice casts \textbf{Infest}. Turing processing occurs: a white or green token dies, a new white or green token is created.
    
    
    \item[T2] Alice casts \textbf{Cleansing Beam}, putting two $+1/+1$ count-ers on the side of the tape we are moving away from.
    
    
    \item[T3] If the Turing machine is remaining in the same state, Alice casts \textbf{Coalition Victory}. If it is changing state, Alice casts \textbf{Soul Snuffers}, putting a \minusminus counter on each creature.
    
    
    \item[T4] If the Turing machine is remaining in the same state, this is the point where Alice casts \textbf{Soul Snuffers}. Otherwise, the next computational step begins.
    
\end{itemize}

\begin{table*}[ht]  
    \caption{Game state when the (2,~18) UTM begins}
    \label{tab:gamestate}
    \centering
    \begin{minipage}[t]{0.15\textwidth} 
    \end{minipage}
    \hfill
    \begin{minipage}[t]{0.5\textwidth}
    \begin{tabular}{l|l|l}
        Card & Controller & Changed text / choices / attachment \\
        \hline
        29 Rotlung Reanimator & Bob & See Table \ref{tab:program} \\
        7 Xathrid Necromancer & Bob & See Table \ref{tab:program} \\
        29 Cloak of Invisibility &Alice& attached to each Rotlung Reanimator \\
        7 Cloak of Invisibility &Alice& attached to each Xathrid Necromancer \\
        Wheel of Sun and Moon  & Alice & attached to Alice\\
        Illusory Gains  & Alice & attached to latest tape token\\
        Steely Resolve & Alice & \textit{Assembly-Worker} \\
        2 Dread of Night & Alice & \textit{Black}\\
        Fungus Sliver & Alice &\textit{Incarnation}\\
        Rotlung Reanimator & Alice & \textit{Lhurgoyf, black, Cephalid}\\
        Rotlung Reanimator & Bob & \textit{Lhurgoyf, green, Lhurgoyf}\\
        Shared Triumph & Alice &\textit{Lhurgoyf}\\
        Rotlung Reanimator & Alice & \textit{Rat, black, Cephalid}\\
        Rotlung Reanimator & Bob & \textit{Rat, white, Rat}\\
        Shared Triumph & Alice &\textit{Rat}\\
    \end{tabular}
    \end{minipage}
    \hfill
    \begin{minipage}[t]{0.3\textwidth}
    \begin{tabular}{l|l}
        Card & Controller \\
        \hline
        Wild Evocation & Bob\\
        Recycle & Bob \\
        Privileged Position & Bob \\
        Vigor & Alice \\
        Vigor & Bob \\
        Mesmeric Orb & Alice \\
        Ancient Tomb & Alice \\
        Prismatic Omen & Alice \\
        Choke & Alice \\
        Blazing Archon & Alice\\
        Blazing Archon & Bob\\
        \\
        \\
        \\
        \\
    \end{tabular}
    \end{minipage}
\end{table*}
\subsection{Beginning a Computational Step and Casting Spells}
At the beginning of a computational step, it is Alice's turn and she has the card \textbf{Infest} in hand. Her library consists of the other cards she will cast during the computation (\textbf{Cleansing Beam}, \textbf{Coalition Victory}, and \textbf{Soul Snuffers}, in that order). Bob's hand and library are both empty. The Turing machine is in its starting state and the tape has already been initialised.

At the start of each of Alice's turns, she has one card in hand. She's forced to cast it due to Bob controlling \textbf{Wild Evocation}, which reads ``At the beginning of each player's upkeep, that player reveals a card at random from their hand. If it's a land card, the player puts it onto the battlefield. Otherwise, the player casts it without paying its mana cost if able.'' When the card resolves, it would normally be put into her graveyard, but Alice is enchanted by \textbf{Wheel of Sun and Moon}, which causes it to be placed at the bottom of her library instead, allowing her to redraw it in the future and keeping the cards she needs to cast in order. After her upkeep step, Alice proceeds to her draw step and draws the card that she will cast next turn.

Alice has no choices throughout this process: she does control one land, but it remains permanently tapped because of \textbf{Choke} (``Islands don't untap during their controllers’ untap steps''), so she is unable to cast any of the spells she draws except via \textbf{Wild Evocation}'s ability.
Neither player is able to attack because they both control a \textbf{Blazing Archon}, ``Creatures can't attack you.''

Bob has no cards in hand and controls \textbf{Recycle}, which reads (in part) ``Skip your draw step''. This prevents Bob from losing due to drawing from an empty library.

\subsection{Reading the Current Cell}
On the first turn of the cycle, Alice is forced to cast \textbf{Infest}, ``All creatures get \minusminustwo until end of turn.'' This kills one creature: the tape token at the position of the current read head, controlled by Bob. This will cause precisely one creature of Bob's to trigger -- either a \textbf{Rotlung Reanimator} or a \textbf{Xathrid Necromancer}. Which precise one triggers is based on that token's creature type and the machine's current state, corresponding to the appropriate rule in the definition of the (2,~18) UTM. This Reanimator or Necromancer will create a new 2/2 token to replace the one that died. The new token's creature type represents the symbol to be written to the current cell, and the new token's colour indicates the direction for the machine to move: white for left or green for right.

Alice controls \textbf{Illusory Gains}, an Aura which reads ``You control enchanted creature. Whenever a creature enters the battlefield under an opponent's control, attach Illusory Gains to that creature.'' Each time one of Bob's \textbf{Rotlung Reanimator}s or \textbf{Xathrid Necromancer}s creates a new token, \textbf{Illusory Gains} triggers, granting Alice control of the newest token on the tape, and reverting control of the previous token to Bob. So at any point Bob controls all of the tape except for the most recently written symbol, which is controlled by Alice.
\begin{table*}[ht]  
\caption{Full text of the Rotlung Reanimators and Xathrid Necromancers encoding the (2,~18) UTM}
\label{tab:program}
\centering\begin{tabular}{cccc|l@{ }l@{ }l@{ }l@{ }l@{ }l@{ }l@{ }l}
\multicolumn{4}{c}{Rogozhin's program} & \multicolumn{6}{l}{Card text} \\
      \hline
$q_1$ &  $1                    $ &  $c_2                  $ &  $Lq_1$       &  Whenever an&  Aetherborn       & dies, create a & &2/2& white&  Sliver  \\
$q_1$ &  $\overrightarrow{1}   $ &  $\overleftarrow{1}_1  $ &  $Rq_1$       &  Whenever a&  Basilisk       & dies, create a & &2/2& green&  Elf    \\
$q_1$ &  $\overleftarrow{1}    $ &  $c_2                  $ &  $Lq_1$       &  Whenever a&  Cephalid     & dies, create a & &2/2& white&  Sliver  \\
$q_1$ &  $\overrightarrow{1}_1 $ &  $1                    $ &  $Rq_1$       &  Whenever a&  Demon      & dies, create a & &2/2& green&  Aetherborn                      \\
$q_1$ &  $\overleftarrow{1}_1  $ &  $\overrightarrow{1}_1 $ &  $Lq_1$       &  Whenever an&  Elf      & dies, create a & &2/2& white&  Demon   \\
$q_1$ &  $b                    $ &  $\overleftarrow{b}    $ &  $Rq_1$       &  Whenever a&  Faerie       & dies, create a & &2/2& green&  Harpy      \\
$q_1$ &  $\overrightarrow{b}   $ &  $\overleftarrow{b}_1  $ &  $Rq_1$       &  Whenever a&  Giant     & dies, create a & &2/2& green&  Juggernaut    \\
$q_1$ &  $\overleftarrow{b}    $ &  $b                    $ &  $Lq_1$       &  Whenever a&  Harpy     & dies, create a & &2/2& white&  Faerie                      \\
$q_1$ &  $\overrightarrow{b}_1 $ &  $b                    $ &  $Rq_1$       &  Whenever an&  Illusion      & dies, create a & &2/2& green&  Faerie                      \\
$q_1$ &  $\overleftarrow{b}_1  $ &  $\overrightarrow{b}_1 $ &  $Lq_1$       &  Whenever a&  Juggernaut       & dies, create a & &2/2& white&  Illusion   \\
$q_1$ &  $b_2                  $ &  $b_3                  $ &  $Lq_2$       &  Whenever a&  Kavu    & dies, create a & tapped&2/2& white&  Leviathan  \\
$q_1$ &  $b_3                  $ &  $\overrightarrow{b}_1 $ &  $Lq_2$       &  Whenever a&  Leviathan    & dies, create a & tapped&2/2& white&  Illusion   \\
$q_1$ &  $c                    $ &  $\overrightarrow{1}   $ &  $Lq_2$       &  Whenever a&  Myr  & dies, create a & tapped&2/2& white&  Basilisk     \\
$q_1$ &  $\overrightarrow{c}   $ &  $\overleftarrow{c}    $ &  $Rq_1$       &  Whenever a&  Noggle  & dies, create a & &2/2& green&  Orc     \\
$q_1$ &  $\overleftarrow{c}    $ &  $\overrightarrow{c}_1 $ &  $Lq_1$       &  Whenever an&  Orc   & dies, create a & &2/2& white&  Pegasus  \\
$q_1$ &  $\overrightarrow{c}_1 $ &  $\overleftarrow{c}_1  $ &  $Rq_2$       &  Whenever a&  Pegasus   & dies, create a & tapped&2/2& green&  Rhino  \\
$q_1$ &  $\overleftarrow{c}_1  $ &  $HALT                 $ &  $    $       &  Whenever a&  Rhino & dies, create a & &2/2& blue& Assassin \\
$q_1$ &  $c_2                  $ &  $\overleftarrow{1}    $ &  $Rq_1$       &  Whenever a&  Sliver  & dies, create a & &2/2& green&  Cephalid      \\
\midrule
$q_2$ &  $1                    $ &  $\overleftarrow{1}   $ &  $Rq_2$        &  Whenever an&  Aetherborn       & dies, create a & &2/2& green&  Cephalid     \\
$q_2$ &  $\overrightarrow{1}   $ &  $\overleftarrow{1}   $ &  $Rq_2$        &  Whenever a&  Basilisk       & dies, create a & &2/2& green&  Cephalid     \\
$q_2$ &  $\overleftarrow{1}    $ &  $\overrightarrow{1}  $ &  $Lq_2$        &  Whenever a&  Cephalid     & dies, create a & &2/2& white&  Basilisk    \\
$q_2$ &  $\overrightarrow{1}_1 $ &  $\overleftarrow{1}_1 $ &  $Rq_2$        &  Whenever a&  Demon      & dies, create a & &2/2& green&  Elf   \\
$q_2$ &  $\overleftarrow{1}_1  $ &  $1                   $ &  $Lq_2$        &  Whenever an&  Elf      & dies, create a & &2/2& white&  Aetherborn                     \\
$q_2$ &  $b                    $ &  $b_2                 $ &  $Rq_1$        &  Whenever a&  Faerie       & dies, create a & tapped&2/2& green&  Kavu                   \\
$q_2$ &  $\overrightarrow{b}   $ &  $\overleftarrow{b}   $ &  $Rq_2$        &  Whenever a&  Giant     & dies, create a & &2/2& green&  Harpy     \\
$q_2$ &  $\overleftarrow{b}    $ &  $\overrightarrow{b}  $ &  $Lq_2$        &  Whenever a&  Harpy     & dies, create a & &2/2& white&  Giant    \\
$q_2$ &  $\overrightarrow{b}_1 $ &  $\overleftarrow{b}_1 $ &  $Rq_2$        &  Whenever an&  Illusion      & dies, create a & &2/2& green&  Juggernaut   \\
$q_2$ &  $\overleftarrow{b}_1  $ &  $\overrightarrow{b}  $ &  $Lq_2$        &  Whenever a&  Juggernaut       & dies, create a & &2/2& white&  Giant    \\
$q_2$ &  $b_2                  $ &  $b                   $ &  $Rq_1$        &  Whenever a&  Kavu    & dies, create a & tapped&2/2& green&  Faerie                     \\
$q_2$ &  $b_3                  $ &  $\overleftarrow{b}_1 $ &  $Rq_2$        &  Whenever a&  Leviathan    & dies, create a & &2/2& green&  Juggernaut   \\
$q_2$ &  $c                    $ &  $\overleftarrow{c}   $ &  $Rq_2$        &  Whenever a&  Myr  & dies, create a & &2/2& green&  Orc    \\
$q_2$ &  $\overrightarrow{c}   $ &  $\overleftarrow{c}   $ &  $Rq_2$        &  Whenever a&  Noggle  & dies, create a & &2/2& green&  Orc    \\
$q_2$ &  $\overleftarrow{c}    $ &  $\overrightarrow{c}  $ &  $Lq_2$        &  Whenever an&  Orc   & dies, create a & &2/2& white&  Noggle   \\
$q_2$ &  $\overrightarrow{c}_1 $ &  $c_2                 $ &  $Rq_2$        &  Whenever a&  Pegasus   & dies, create a & &2/2& green&  Sliver                   \\
$q_2$ &  $\overleftarrow{c}_1  $ &  $c_2                 $ &  $Lq_1$        &  Whenever a&  Rhino & dies, create a & tapped&2/2& white&  Sliver                   \\
$q_2$ &  $c_2                  $ &  $c                   $ &  $Lq_2$        &  Whenever a&  Sliver  & dies, create a & &2/2& white&  Myr
\end{tabular}
\end{table*}
\subsection{Moving Left or Right}
If the new token is white, the Turing machine needs to move left. To do this we need to take two actions: put a $+1/+1$ counter on all white creatures (move the tape away from white), and put a \minusminus counter on all green creatures (move the tape towards green). We rephrase this instead as: put two $+1/+1$ counters on all white creatures, and put a \minusminus counter on \textit{all} creatures.

On Alice's second turn, she casts \textbf{Cleansing Beam}, which reads ``Cleansing Beam deals 2 damage to target creature and each other creature that shares a color with it.'' Bob controls \textbf{Privileged Position} so none of Bob's creatures can be targeted by any spell Alice casts. Alice controls some creatures other than the tape token, but they have all been granted creature type Assembly-Worker by a hacked \textbf{Olivia Voldaren}, and Alice controls a \textbf{Steely Resolve} naming Assembly-Worker (``Creatures of the chosen type have shroud. \textit{(They can't be the targets of spells or abilities.)}'') This makes it so that the only legal target for \textbf{Cleansing Beam} is the one tape token that Alice controls thanks to her \textbf{Illusory Gains}. 

Recall that this token is white if we're moving left, or green if we're moving right. \textbf{Cleansing Beam} is about to deal 2 damage to each white creature if we're moving left, or to each green creature if we're moving right. Alice and Bob each control a copy of \textbf{Vigor} -- ``If damage would be dealt to another creature you control, prevent that damage. Put a $+1/+1$ counter on that creature for each 1 damage prevented this way.'' So \textbf{Cleansing Beam} ends up putting two $+1/+1$ counters on either each white creature or each green creature. 

On the last turn of the cycle, Alice casts \textbf{Soul Snuffers}, a 3/3 black creature which reads ``When Soul Snuffers enters the battlefield, put a \minusminus counter on each creature.'' There are two copies of \textbf{Dread of Night} hacked to each say ``\textit{Black} creatures get \minusminusnospace'', which mean that the \textbf{Soul Snuffers}' triggered ability will kill itself, as well as shrinking every other creature. The creatures comprising the tape have now received either a single \minusminus counter, or two $+1/+1$ counters and a \minusminus counter.

To ensure that the creatures providing the infrastructure (such as \textbf{Rotlung Reanimator}) aren't killed by the succession of \minusminus counters each computational step, we arrange that they also have game colours green, white, red and black, using \textbf{Prismatic Lace}, ``Target permanent becomes the color or colors of your choice. \textit{(This effect lasts indefinitely.)}'' Accordingly, each cycle \textbf{Cleansing Beam} will put two $+1/+1$ counters on them, growing them faster than the \minusminus counters shrink them. This applies to each creature except \textbf{Vigor} itself; to keep each player's \textbf{Vigor} from dwindling, there is a \textbf{Fungus Sliver} hacked to read ``All \textit{Incarnation} creatures have `Whenever this creature is dealt damage, put a $+1/+1$ counter on it.' ''

\subsection{Changing State}
The instruction to change state is handled by replacing seven of Bob's \textbf{Rotlung Reanimator}s with \textbf{Xathrid Necromancer}. These two cards have very similar text, except that \textbf{Xathrid Necromancer} only notices Bob's creatures dying (this is not a problem, as the active cell of the tape is always controlled by Bob), and that \textbf{Xathrid Necromancer} creates its token \textit{tapped}.

For example, when the $q_1$ program (State A) sees symbol 1, it writes symbol 18, moves left, and remains in state A. This is represented by a phasing \textbf{Rotlung Reanimator} under Bob's control saying ``Whenever Rotlung Reanimator or another \textit{Aetherborn} dies, create a 2/2 \textit{white} \textit{Sliver} creature token.''

By contrast, when the $q_1$ program sees symbol 11, it writes symbol 12, moves left, and changes to state B. This is represented by a phasing \textbf{Xathrid Necromancer} under Bob's control saying ``Whenever Xathrid Necromancer or another \textit{Kavu} creature you control dies, create a tapped 2/2 \textit{white} \textit{Leviathan} creature token.''

In both cases this token is created under Bob's control on turn T1, but Alice's \textbf{Illusory Gains} triggers and grants her control of it. In the case where it's tapped, that means at the beginning of turn T2, it will untap. This causes Alice's \textbf{Mesmeric Orb}'s trigger to be put on the stack at the same time as Bob's \textbf{Wild Evocation}'s trigger (since no player receives priority during the untap step). Alice is the active player, so Alice's trigger is put on the stack first and then Bob's\cite{rules}; so the \textbf{Wild Evocation} trigger resolves, forcing Alice to cast and resolve \textbf{Cleansing Beam}, before the \textbf{Mesmeric Orb} trigger resolves. 

When the \textbf{Mesmeric Orb} trigger does resolve, it tries to put the \textbf{Coalition Victory} from the top of Alice's library into her graveyard. But \textbf{Wheel of Sun and Moon} modifies this event to put \textbf{Coalition Victory} onto the bottom of her library, just underneath the \textbf{Cleansing Beam} that's just resolved.

Once all these triggers are resolved, Alice proceeds to her draw step. When the state is not changing, she will draw \textbf{Coalition Victory} at this point, but when the state is changing, that card is skipped and she moves on to draw \textbf{Soul Snuffers} in turn T2's draw step, so she will cast it on turn T3.

The net result of this is that the computation step is 3 turns long for each player when the state is changing, but 4 turns long for each player when the state is not changing. In the normal 4-turn operation, Bob's phasing Reanimators and Necromancers will phase in twice and phase out twice, and be in the same state on one cycle's turn T1 as they were in the previous cycle's turn T1. But when changing state, they will have changed phase by the next cycle's turn T1, switching the Turing machine's state.

\subsection{Out of Tape}
The Turing tape can be initialised to any desired length before starting processing. But it is preferable to allow the machine to run on a simulated infinite tape: in other words, to assume that any uninitialised tape space contains symbol 3 (the blank symbol in the (2,~18) UTM), represented by creature type Cephalid. This is accomplished by having the ends of the currently-initialised tape marked by two special tokens, one green Lhurgoyf and one white Rat. 
Suppose we've exhausted all the initialised tape to the left. This means that the casting of \textbf{Infest} on turn T1 kills the Lhurgoyf rather than one of the normal tape types. This does not directly trigger any of the normal Reanimators/Necromancers. Instead, Bob has another \textbf{Rotlung Reanimator} hacked to read ``Whenever Rotlung Reanimator or another \textit{Lhurgoyf} dies, create a 2/2 \textit{green} \textit{Lhurgoyf} creature token'', and Alice has a \textbf{Rotlung Reanimator} hacked to read ``Whenever Rotlung Reanimator or another \textit{Lhurgoyf} dies, create a 2/2 \textit{black} \textit{Cephalid} creature token.'' Bob's trigger will resolve first, then Alice's.

First, Bob's Reanimator trigger creates a new Lhurgoyf just to the left of the current head. (Alice's \textbf{Illusory Gains} triggers and gives her control of this new Lhurgoyf, but that will soon change.) We have one copy of \textbf{Shared Triumph} set to Lhurgoyf (``Creatures of the chosen type get +1/+1'') so this token arrives as a 3/3.

Second, Alice's Reanimator trigger now creates a 2/2 black Cephalid under Alice's control. The same two copies of \textbf{Dread of Night} as before are giving all black creatures \minusminustwonospace, so the black Cephalid will arrive as a 0/0 and immediately die.

The death of this Cephalid triggers one of the regular phasing Reanimators of Bob's just as if a tape cell containing symbol 3 had been read: a new 2/2 token is created and \textbf{Illusory Gains} moves to that new token. The green Lhurgoyf token serving as an end-of-tape marker has been recreated one step over to the left.

The situation for the white Rat representing the right-hand end of the tape is exactly equivalent. Bob has a \textbf{Rotlung Reanimator} hacked to read ``Whenever Rotlung Reanimator or another \textit{Rat} dies, create a 2/2 \textit{white} \textit{Rat} creature token''; Alice has a \textbf{Rotlung Reanimator} hacked to read ``Whenever Rotlung Reanimator or another \textit{Rat} dies, create a 2/2 \textit{black} \textit{Cephalid} creature token''; and we have another \textbf{Shared Triumph} set to Rat.

(This algorithm would be a little more complex if reading symbol 3 could cause a state change in the (2,~18) UTM, but thankfully it cannot.)

\subsection{Halting}
We choose to encode halting as making Alice win the game.

When the Turing machine doesn't change state, Alice casts the card \textbf{Coalition Victory} on her third turn. It reads ``You win the game if you control a land of each basic land type and a creature of each color.'' This normally accomplishes nothing because she controls no blue creatures (\textbf{Prismatic Lace} has been used to give her creatures of all the other colours). She does, however, control one land, and also controls \textbf{Prismatic Omen}, which reads ``Lands you control are every basic land type in addition to their other types.'' Recall that \textbf{Choke} is in play, preventing her from activating the mana ability of this land.

When the halt symbol is read (symbol 17 in state A), the appropriate phasing Reanimator of Bob's reads ``Whenever Rotlung Reanimator or another \textit{Rhino} dies, create a 2/2 \textit{blue} \textit{Assassin} creature token.'' Alice's \textbf{Illusory Gains} takes control of this Assassin token in the usual way in turn T1. She now meets the condition for \textbf{Coalition Victory} when she casts it on turn T3, and wins the game.

If the encoded machine does not in fact halt then the game has entered an unbreakable deterministic infinite loop, which is specified as a draw by rule 104.4b\cite{rules}.

%


\begin{table*}[hb]  
    \centering
    \caption{60-Card Decklist to play the Turing machine in a Legacy tournament}
    \label{tab:decklist}
    \begin{tabular}{l|l ||l|l ||l|l}
        \ \ Card & Purpose & \ \ Card & Purpose & \ \ Card & Purpose \\
    \hline
4\ \  Ancient Tomb	&Bootstrap&	1\ \  Rotlung Reanimator	&Logic processing&	1\ \  Xathrid Necromancer	&Change state	\\
4\ \  Lotus Petal	&Bootstrap&	1\ \  Cloak of Invisibility	&Logic processing&	1\ \  Mesmeric Orb	&Change state	\\
4\ \  Grim Monolith	&Infinite mana device&	1\ \  Infest	&Logic processing&	1\ \  Coalition Victory	&Halting device	\\
4\ \  Power Artifact	&Infinite mana device&	1\ \  Cleansing Beam	&Logic processing&	1\ \  Prismatic Omen	&Halting device	\\
4\ \  Gemstone Array	&Infinite mana device&	1\ \  Soul Snuffers	&Logic processing&	1\ \  Choke	&Halting device	\\
4\ \  Staff of Domination	&Draw rest of deck&	1\ \  Illusory Gains	&Logic processing&	1\ \  Recycle	&Remove choices	\\
1\ \  Memnarch	&Make token copies&	1\ \  Privileged Position	&Logic processing&	1\ \  Blazing Archon	&Remove choices	\\
1\ \  Stolen Identity	&Make token copies&	1\ \  Steely Resolve	&Logic processing&	1\ \  Djinn Illuminatus	&Simplify setup	\\
1\ \  Artificial Evolution	&Edit cards&	1\ \  Vigor	&Logic processing&	1\ \  Reito Lantern	&Simplify setup	\\
1\ \  Olivia Voldaren	&Edit cards&	1\ \  Fungus Sliver	&Logic processing&	1\ \  Claws of Gix	&Simplify setup	\\
1\ \  Glamerdye	&Edit cards&	1\ \  Dread of Night	&Logic processing&	1\ \  Riptide Replicator	&Set up tape	\\
1\ \  Prismatic Lace	&Edit cards&	1\ \  Wild Evocation	&Forced play device&	1\ \  Capsize	&Set up tape	\\
1\ \  Donate	&Edit card control&	1\ \  Wheel of Sun and Moon	&Forced play device&	1\ \  Karn Liberated	&Cleanup after setup	\\
1\ \  Reality Ripple	&Edit card phase&	1\ \  Shared Triumph	&Infinite tape device&	1\ \  Fathom Feeder	&Cleanup after setup	\\
    \end{tabular}
\end{table*}

\section{Discussion}\label{discussion}

\subsection{Consequences for Computational Theories of Games}\label{theory-of-gaming}

This construction establishes that \textit{Magic: The Gathering} is the most computationally complex real-world game known in the literature. In addition to showing that optimal strategic play in \textit{Magic} is non-computable, it also shows that merely evaluating the deterministic consequences of past moves in \textit{Magic} is non-computable. The full complexity of optimal strategic play remains an open question, as do many other computational aspects of \textit{Magic}. For example, a player appears to have infinitely many moves available to them from some board states of \textit{Magic}. Whether or not there exists a real-world game of \textit{Magic} in which a player has infinitely many \textit{meaningfully different} moves available to them has the potentially to highly impact the way we understand and model games as a form of computation.

Indeed, this result raises several interesting philosophical questions about games as a form of computation. Some authors, such as Demaine and Hearn \cite{dh:contraint-logic}, have sought a formal framework for modelling games that is strictly sub-Turing. Unlike the open-world, non-strategic games in which Turing machines have been constructed before, \textit{Magic: The Gathering} is unambiguously a two-player strategic game like such models attempt to represent. Therefore this result shows that any sub-Turing model is necessarily inadequate to capture all games. Quite the opposite: it seems likely that a \textit{super-Turing} model of games would be necessary to explain \textit{Magic}. The na{\"i}ve extension of Demaine and Hearn's Constraint Logic to allow for unbounded memory appears to be meaningless, although it's possible that a clever approach would bring success.

\begin{open}
Does there exist a generalisation of Constraint Logic that explains the computational complexity of \textit{Magic: The Gathering}?
\end{open}

Although our construction is reducible to the Halting Problem, the fact that even evaluating a board is non-computable is strongly suggestive that the complexity of strategic play is greater than that. We believe there is strong evidence that the true computational complexity is far higher. In particular, we conjecture:

\begin{conj}
Playing \textit{Magic: The Gathering} optimally is at least as hard as $0^{(\omega)}$.
\end{conj}

Whether or not it is possible for there to be a real-world game whose computational complexity is strictly harder than $0^{(\omega)}$ is an interesting philosophical question. If not, then this conjecture would imply that \textit{Magic: The Gathering} is as hard as it is possible for a real-world game to be.

\subsection{Real-world Playability and Legality}

While there are practical difficulties involved with correctly setting up the necessary board state, such as running out of space on your table, a sufficiently tenacious player could set up and execute this construction in a real-world tournament game of \textit{Magic: The Gathering}. An example 60-card deck that is capable of executing this construction on the first turn of the game and which is legal in the competitive Legacy format can be seen in Table \ref{tab:decklist}.

With the correct draw, the deck uses \textbf{Ancient Tomb} and three \textbf{Lotus Petal}s to play \textbf{Grim Monolith} and \textbf{Power Artifact} and generate unlimited colourless mana, at which point \textbf{Staff of Domination} draws the rest of the deck and \textbf{Gemstone Array} generates unlimited coloured mana. The deck casts most of the permanents immediately, and uses \textbf{Stolen Identity} to make token copies of them (using \textbf{Memnarch} first on the enchantments like \textbf{Cloak of Invisibility}). The tape is initialised with \textbf{Riptide Replicator} and \textbf{Capsize}. \textbf{Djinn Illuminatus} or \textbf{Reito Lantern} allow repeated casting of the text-modification cards, as well as \textbf{Reality Ripple} which sets the phase of the \textbf{Rotlung Reanimator}s and \textbf{Donate} which gives most permanents to Bob. Once everything is set up, \textbf{Steely Resolve} is cast, and then \textbf{Karn Liberated} and \textbf{Capsize} are used to exile all setup permanents and all cards from Bob's hand, eventually exiling \textbf{Capsize} and \textbf{Karn Liberated} themselves. Now no player has any remaining choices except to let the Turing machine execute.

In addition to the \textit{Comprehensive Rules} \cite{rules}, play at sanctioned \textit{Magic: The Gathering} tournaments is also governed by the \textit{Tournament Rules} \cite{tournament}. Some of these rules, most notably the ones involving slow play, may effect an individual's ability to successfully execute the combo due to concerns about the sheer amount of time it would take to manually move the tokens around to simulate a computation on a Turing machine. This would not be a concern for two agents with sufficiently high computational power, as the Tournament Rules also provide a mechanism called ``shortcuts'' for players to skip carrying out laborious loops if both players agree on the game state at the beginning and the end of the shortcut.

\section{Conclusion}

We have presented a methodology for embedding Rogozhin's (2,~18) universal Turing machine in a two-player game of \textit{Magic: The Gathering}. Consequently, we have shown that identifying the outcome of a game of \textit{Magic} in which all moves are forced for the rest of the game is undecidable. In addition to solving a decade-old outstanding open problem, in the process of arriving at our result we showed that \textit{Magic: The Gathering} does not fit assumptions commonly made by computer scientists while modelling games. We conjecture that optimal play in \textit{Magic} is far harder than this result alone implies, and leave the true complexity of \textit{Magic} and the reconciliation of \textit{Magic} with existing theories of games for future research.

\section*{Acknowledgements}

We are grateful to C-Y. Howe for help simplifying our Turing machine construction considerably and to Adam Yedidia for conversations about the design and construction of Turing machines.
\bibliographystyle{plain}
\bibliography{citations}

\begin{thebibliography}{10}

\bibitem{at:frontier}
David Auger and Oliver Teytaud.
\newblock The frontier of decidability in partially observable recursive games.
\newblock {\em International Journal of Foundations of Computer Science}, 2012.

\bibitem{bk:biggest}
Stella Biderman and Bj{\o}rn Kjos-Hanssen.
\newblock Non-comparable natural numbers.
\newblock Theoretical Computer Science Stack Exchange, 2018.
\newblock \url{https://cstheory.stackexchange.com/q/41384 (version:
  2018-08-16)}.

\bibitem{ci:blocking}
Krishnendu Chatterjee and Rasmus Ibsen-Jensen.
\newblock The complexity of deciding legality of a single step of
  \textrm{Magic: The Gathering}.
\newblock In {\em 22nd European Conference on Artificial Intelligence}, 2016.

\bibitem{c:old-tm}
Alex Churchill.
\newblock \textit{Magic: The Gathering} is \textrm{Turing} complete v5, 2012.
\newblock \url{https://www.toothycat.net/~hologram/Turing/}.

\bibitem{c:forum}
Alex Churchill et~al.
\newblock \textrm{Magic is Turing} complete (the \textrm{Turing} machine
  combo), 2014.
\newblock \url{http://tinyurl.com/pv3n2lg}.

\bibitem{wcp:mcmc}
Peter I.~Cowling Colin D.~Ward and Edward~J. Powley.
\newblock Ensemble determinization in \textrm{Monte Carlo} tree search for the
  imperfect information card game \textrm{Magic: The Gathering}.
\newblock In {\em IEEE Transactions on Computational Intelligence and AI in
  Games}, volume~4, 2012.

\bibitem{cl:smash}
Michael~J. Coulombe and Jayson Lynch.
\newblock Cooperating in video games? \textrm{Impossible! Undecidability} of
  team multiplayer games.
\newblock In {\em 9th International Conference on Fun with Algorithms}, 2018.

\bibitem{dh:cgt}
Erik~D. Demaine and Robert~A. Hearn.
\newblock Playing games with algorithms: algorithmic combinatorial game theory.
\newblock In {\em 26th Symp. on Mathematical Foundations in Computer Science},
  pages 18--32, 2001.

\bibitem{dh:contraint-logic}
Erik~D. Demaine and Robert~A. Hearn.
\newblock Constraint logic: A uniform framework for modeling computation as
  games.
\newblock In {\em 2008 23rd Annual IEEE Conference on Computational
  Complexity}, pages 149--162, 2008.

\bibitem{dh:games-puzzles-computation}
Erik~D. Demaine and Robert~A. Hearn.
\newblock {\em Games, Puzzles, and Computation}.
\newblock CRC Press, 2009.

\bibitem{e:thesis}
Alexander Esche.
\newblock {\em Mathematical Programming and \textrm{Magic: The Gathering}}.
\newblock PhD thesis, Northern Illinois University, 2018.

\bibitem{f:incomparable}
Eugenio Fortanely.
\newblock Personal communication, 2018.

\bibitem{rice}
H.~G. Rice.
\newblock Classes of recursively enumerable sets and their decision problems.
\newblock {\em Trans. Amer. Math. Soc.}, 74:358–366, 1953.

\bibitem{r:utms}
Yurii Rogozhin.
\newblock Small universal \textrm{Turing} machines.
\newblock {\em Theoretical Computer Science}, 168(2):215--240, 1996.

\bibitem{wc:mcmc}
Colin~D. Ward and Peter~I. Cowling.
\newblock \textrm{Monte Carlo} search applied to card selection in
  \textrm{Magic: The Gathering}.
\newblock In {\em CIG'09 Proceedings of the 5th international conference on
  Computational Intelligence and Games}, pages 9--16, 2009.

\bibitem{rules}
{Wizards of the Coast}.
\newblock \textrm{Magic: The Gathering} comprehensive rules, Aug 2018.
\newblock
  \url{https://magic.wizards.com/en/game-info/gameplay/rules-and-formats/rules}.

\bibitem{tournament}
{Wizards of the Coast}.
\newblock \textrm{Magic: The Gathering} tournament rules, Aug 2018.
\newblock
  \url{https://wpn.wizards.com/sites/wpn/files/attachements/mtg_mtr_21jan19_en.pdf}.

\end{thebibliography}

\end{document}